\documentclass[preprint,12pt,authoryear]{elsarticle}
\usepackage[colorlinks,linkcolor=blue]{hyperref}
\usepackage{natbib,lineno,hyperref,amsmath,graphicx,epstopdf,epsfig,multirow,booktabs,subfigure,amsfonts,amssymb}
\modulolinenumbers[5]
\usepackage{multicol}
\journal{ISPRS Journal of Photogrammetry and Remote Sensing}
\usepackage{geometry}
\biboptions{sort}

\begin{document}

\begin{frontmatter}

  \title{Attention Aware Cost Volume Pyramid Based Multi-view Stereo Network for 3D Reconstruction}


  \author{Anzhu Yu \corref{coauthor} \corref{mycorrespondingauthor}}
  \ead{anzhu\_yu@126.com}
  \author{Wenyue Guo \corref{coauthor}}
  \ead{guowyer@163.com}
  \author{Bing Liu \corref{coauthor}}
  \ead{liubing220524@126.com}

  \author{Xin Chen}
  \ead{xinchen\_cosmos@126.com}
  \author{Xin Wang}
  \ead{rainbowwxdo@163.com}
  \author{Xuefeng Cao}
  \ead{CAO\_Xue\_Feng@163.com}
  \author{Bingchuan Jiang}
  \ead{jbc021@163.com}

  \cortext[coauthor]{Prof. A.Yu, Dr. B. Liu and Dr. W. Guo have equal contribution to this work and are co-first authors.}
  \cortext[mycorrespondingauthor]{Corresponding author: Anzhu Yu.}

  \address{PLA Strategic Support Force Information Engineering University, Zhengzhou, 450000,China}

  \begin{abstract}
    We present an efficient multi-view stereo (MVS) network for
    3D reconstruction from multi-view images. While previous learning based reconstruction approaches performed quite well, 
    most of them estimate depth maps at a fixed resolution using plane sweep volumes with a fixed depth hypothesis at each plane, which requires densely sampled planes for desired accuracy and therefore is difficult to achieve high resolution depth maps.
    In this paper we introduce a coarse-to-fine depth inference strategy to achieve high resolution depth.
    This strategy estimates the depth map at coarsest level, while the depth maps at finer levels are considered as the upsampled depth map from previous level with pixel-wise depth residual. Thus, we
    narrow the depth searching range with priori information from previous level and construct new cost volumes from the pixel-wise depth residual to perform depth map refinement. Then the final depth map could be achieved iteratively since all the parameters are shared between different levels.
    At each level, the self-attention layer is introduced to the feature extraction block for capturing the long range dependencies for depth inference task, and the cost volume is generated using similarity measurement instead of the variance based methods used in previous work.
    Experiments were conducted on both the DTU benchmark dataset and recently released BlendedMVS dataset. The results demonstrated that our model could
    outperform most state-of-the-arts (SOTA) methods. The codebase of this project is at \href{https://github.com/ArthasMil/AACVP-MVSNet}{https://github.com/ArthasMil/AACVP-MVSNet}.

  \end{abstract}

\end{frontmatter}

\section{Introduction}

Multi-view stereo targets at reconstructing the observed 3D scene with dense representation from multi-view images and corresponding camera parameters,
which has been extensively studied for decades and covers a wide range of applications such as photogrammetry \citep{rottensteiner2014results,malihi20163d,masiero20193d}, cartography \citep{bitelli2018integrated,buyukdemircioglu2020reconstruction}
and augmented reality \citep{yang2013image,sing2016garden,harazono2019development}. MVS also remains a core problem of photogrammetry and computer vision in many aspects
\citep{hirschmuller2007stereo,tola2012efficient,zhu2017deep,zbontar2015computing,yao2018mvsnet}.

The traditional MVS methods introduce the hand-crafted similarity metrics for image association followed by
some optimization steps for dense point cloud generation (e.g., normalized cross correlation as similarity metric and semi-global matching for optimization \citep{hirschmuller2007stereo}). Though these
methods perform quite well under ideal
Lambertian scenarios with high textured regions, they suffer from incomplete reconstruction in some low-textured, reflective regions where the accuracy and robustness of dense matching decrease \citep{yao2018mvsnet}. 
In the meantime, the traditional
methods are usually conducted sequentially, which is a time and memory consuming procedure and limits its application in situation where efficiency is needed. 
The Simultaneous Localization And Mapping (SLAM) is also a popular way for sensing the 3D structure of surroundings.
Though SLAM could extract a lot of 2D/3D landmarks (A.K.A the object points in photogrammetry application) along
with the transformation matrix of each frame, the density of generated point cloud is usually not enough for 3D reconstruction for the reason that 3D reconstruction usually 
 needs pixel-wise matching and geo-positioning instead of 3D point cloud of sparse landmarks. Overall, accurate and efficient 3D reconstruction is still a hot and challenging topic.

Owing to rapid developments of the Deep Learning technique \citep{lecun2015deep}, some learning based MVS models arise in recent years \citep{hartmann2017learned,ji2017surfacenet,kar2017learning,xiang2020pruning}. The convolutional neural networks (CNNs) that could extract hierarchical features with stronger
representation ability make it possible to speed up MVS processing by using the computation power of GPUs. Some end-to-end learning based MVS methods were presented by modeling the regression relationship between multi-view images and corresponding
depth maps \citep{chen2019point,yao2018mvsnet,xiang2020pruning}, after which the estimated depth maps could be fused and processed to generate the dense 3D point clouds of target region (e.g., using the Fusibile toolbox \citep{galliani2015massively}). Though the aforementioned literatures could achieve better 
results in terms of accuracy and completeness compared to traditional MVS methods, they require huge GPU memory caused by high dimension cost volume \citep{esteban2004silhouette,seitz2006comparison} for depth inference. A normal way to release memory burden is to inference depth map with downsampled images \citep{yao2018mvsnet,yao2019recurrent}. 
It comes at the cost of reconstruction accuracy and completeness, however.

To this end, we endeavor to propose an Attention Aware Cost Volume Pyramid Multi-view Stereo Network (AACVP-MVSNet) for 3D reconstruction. In summary, our main contributions are listed below: 
\begin{itemize}
  \item We introduce the self-attention layers for hierarchical features extraction, which may capture important information for the subsequent depth inference task.
  \item We introduce the similarity measurement for cost volume generation instead of the variance based methods used by most MVS networks.
  \item We use a coarse-to-fine strategy for depth inference which is applicable for large scaled images, and extensive experiments validate that the proposed approach achieves an overall performance improvement than most SOTA algorithms on DTU dataset.
\end{itemize}

\section{Related work}
\subsection{Traditional MVS methods}
In 3D reconstruction workflow, the traditional MVS methods are usually implemented following the sparse point cloud generation
procedure \citep{ma2018review} (e.g. the Structure from Motion computation (SfM) \citep{schonberger2016structure,yang2013image}). To reconstruct the
dense 3D point cloud, the recovered intrinsic and extrinsic parameters of cameras for each image and the sparse point cloud obtained from SfM or SLAM
are set as inputs. The Clustering Views for Multi-view Stereo (CMVS) \citep{2010Towards} and the Patch-based Multi-view Stereo (PMVS) \citep{Koch2014Achievements} are 
very popular methods for dense 3D reconstruction. The CMVS introduces SfM filter to merge extracted feature points and decomposes the input images into 
a set of image clusters of manageable size,
after which the MVS software could be used for 3D reconstruction. The PMVS uses clustered images from CMVS as input and generates dense
3D point clouds through matching, expansion and filtering.

The Semi-Global Matching (SGM) is also widely way for 3D reconstruction, which is proposed for estimation of a dense disparity map from rectified stereo image pairs with introducing the penalty of inconsistency\citep{Heiko2005Accurate}.
As the SGM algorithm has trade-off between computing time and the quality of results, it's faster than PMVS and has encountered wide adoption in real-time stereo vision applications \citep{hirschmuller2007stereo}.

Though the aforementioned works yield impressive results and perform well on the accuracy, they require photometric consistency and 
would achieve unsatisfactory matching results with hand-crafted features and similarity when dealing with non-Lambertian surfaces, low textured and texture-less regions \citep{xiang2020pruning}.
Therefore, the traditional MVS methods still need to be improved to achieve more robust and complete reconstruction results \citep{galliani2015massively, vu2011high}. 

\subsection{Learned stereo algorithms}
In contrast to traditional stereo matching algorithms used in traditional MVS methods that introduce hand-crafted image features and
matching metrics \citep{hirschmuller2007evaluation}, the learned stereo algorithms introduce deep learning
networks to achieve better matching results.

Most of learned stereo algorithms focus on image association procedures, in which the Convolutional Neural Networks (CNNs)
are used for hierarchical features extraction and matching. \cite{luo2016efficient} and \cite{vzbontar2016stereo} use CNNs to
extract learned features for matching, followed by the SGM for dense reconstruction. The SGMNet \citep{seki2017sgm} introduces CNNs for
penalty-parameters estimation and outperformed state-of-the-art accuracy on KITTI benchmark datasets. \cite{zhang2019deep} introduces Graph Neural Network (GNN) model for feature matching, which transforms coordinates of
feature points into local features to replace NP-hard assignment problem in some traditional methods.
The CNN-CRF introduces conditional random field into networks and forms an end-to-end matching algorithm \citep{knobelreiter2017end}.
The RF-Net \citep{shen2019rf} is an end-to-end trainable matching network based on receptive field to compute sparse correspondence between images. In the work of \cite{kendall2017end}, a deep learning architecture is proposed
for regressing disparity from a rectified pair of stereo images and the cost volume is used for feature aggregation.

\subsection{Learned MVS algorithms}
There are few deep learning algorithms for MVS problem before the work of \cite{hartmann2017learned}. The SurfaceNet \citep{ji2017surfacenet} is the pioneer to build a learning based pipeline for MVS problem, which builds the cost volume
with sophisticated voxel-wise view selection and
use 3D regularization for inferencing surface voxels. \cite{yao2018mvsnet} proposes MVSNet for the MVS problem, which introduces differentiable homography to build the cost volume
for features aggregation and use 3D regularization for depth inference. To reduce the memory burden, \cite{yao2019recurrent} proposes R-MVSNet, which sequentially regularized the 2D cost
maps along the depth direction via the gated recurrent
unit (GRU).  Based on these works, many learned MVS networks are proposed to improve the accuracy and completeness of the 3D reconstruction results, such as the
PointMVSNet \citep{ChenPMVSNet2019ICCV}, the CasMVSNet \citep{gu2020cascade}, the PVA-MVSNet \citep{yi2020PVAMVSNET} and the CVP-MVSNet \citep{yang2020cost}. Some of them use a coarse-to-fine strategy to improve
the accuracy and completeness of reconstruction results compared to the of original MVSNet, as the depth map could be estimated with higher resolution input images whereby better 3D reconstruction would be achieved.

Feature extraction is a key issue for learned MVS algorithms. Regarding another key issue, cost volume generation, though the aforementioned literatures introduce CNN blocks for feature extraction, 
they hardly capture long-range dependencies during coarse-to-fine strategy and fail to capture the important information for depth inference tasks.
Though the variance based cost metric is widely used for cost volume generation, it is pointed out by \cite{tulyakov2018practical} that the number of channels of the cost volume could be reduced and similar accuracy can still be achieved, 
which implied that the variance based cost volume with a lot of channels may be redundant and the memory consumption as well as the computational requirement could be reduced.
To overcome and address these issues, we introduce the self-attention mechanism into depth inference procedure to improve the overall accuracy of reconstruction results, 
and implement a similarity based method for cost volume generation to simultaneously reduce the computational requirement and the memory consumption.

\section{Methodology}
This section describes the detailed architecture of the proposed network. The design of AACVP-MVSNet strongly
borrows the insights from previous MVS approaches and novel feature extraction methods.

\begin{figure}[h]
  \begin{center}
    \includegraphics[scale = 0.130]{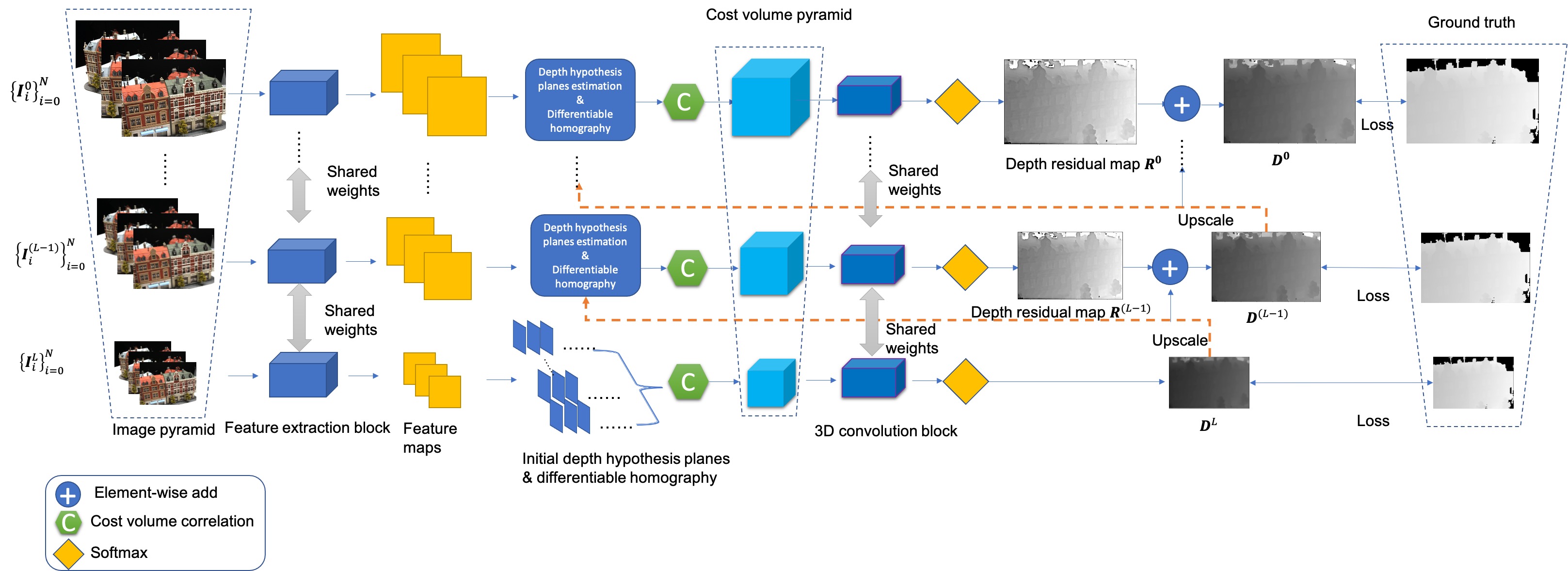}
    \caption{The network structure of AACVP-MVSNet. The feature extraction block and the 3D convolution block are both weights-shared 
    between all levels. The image pyramid is built firstly, and the iterative depth estimation starts at the coarsest level. The depth map estimated at each level is taken 
    as input at next level for depth residual estimation.}\label{fig:NetworkStructure}
  \end{center}
\end{figure}

The overall system is depicted in Fig.(\ref{fig:NetworkStructure}). The multi-view images are first downsampled to form image pyramid, after which a weights-shared 
feature extraction block is built for feature extraction at each level. The depth inference begins at coarse level (level $L$) by building the cost volume $\mathbf{C}^L$ using the similarity measurement,
namely the cost volume correlation which uses the similarity metrics rather than variance based metrics at bottom in Fig.(\ref{fig:NetworkStructure}), and the initial depth map is generated by cost volume regularization using the 3D convolution block and 
the softmax operation. The estimated depth map $\mathbf{D}^L$ is upscaled to the image size of next level (level $L-1$) and the cost volume $\mathbf{C}^{(L-1)}$ is built with depth hypothesis planes 
estimation followed by cost volume correlation. 
The depth residual map $\mathbf{R}^{(L-1)}$ is also estimated with 3D convolution block and the softmax operation, and the depth map $\mathbf{D}^{(L-1)}$ is upscaled to the image size 
of the level $L-2$ for depth inference at $(L-2)$th level. Thus, an iterative depth map estimation procedure is formed along with a cost volume pyramid $\{\mathbf{C}^i\}(i = L,L-1,\cdots, 0)$.

As existing works, we assume the reference image is denoted as $\mathbf{I}_0 \in \mathbb{R}^{H \times W}$, where $H$
and $W$ are the height and width of the input image respectively. Let $\{\mathbf{I_i}\}_{i=1}^N$ be input the $N$ source images for reconstruction. For MVS problem,
the corresponding camera intrinsic matrix, rotation matrix, and translation vector denoted as $\{\mathbf{K}_i,\mathbf{R}_i, \mathbf{t}_i \}_{i=0}^N$
are known for all input views. Our goal is to estimate the depth map $\mathbf{D}^0$ from $\{\mathbf{I_i}\}_{i=0}^N$ with given $\{\mathbf{K}_i,\mathbf{R}_i, \mathbf{t}_i \}_{i=0}^N$.

\subsection{Self-attention based hierarchical feature extraction}
\subsubsection{Self-attention based feature extraction block}
\begin{figure}[h]
  \begin{center}
    \includegraphics[scale = 0.5]{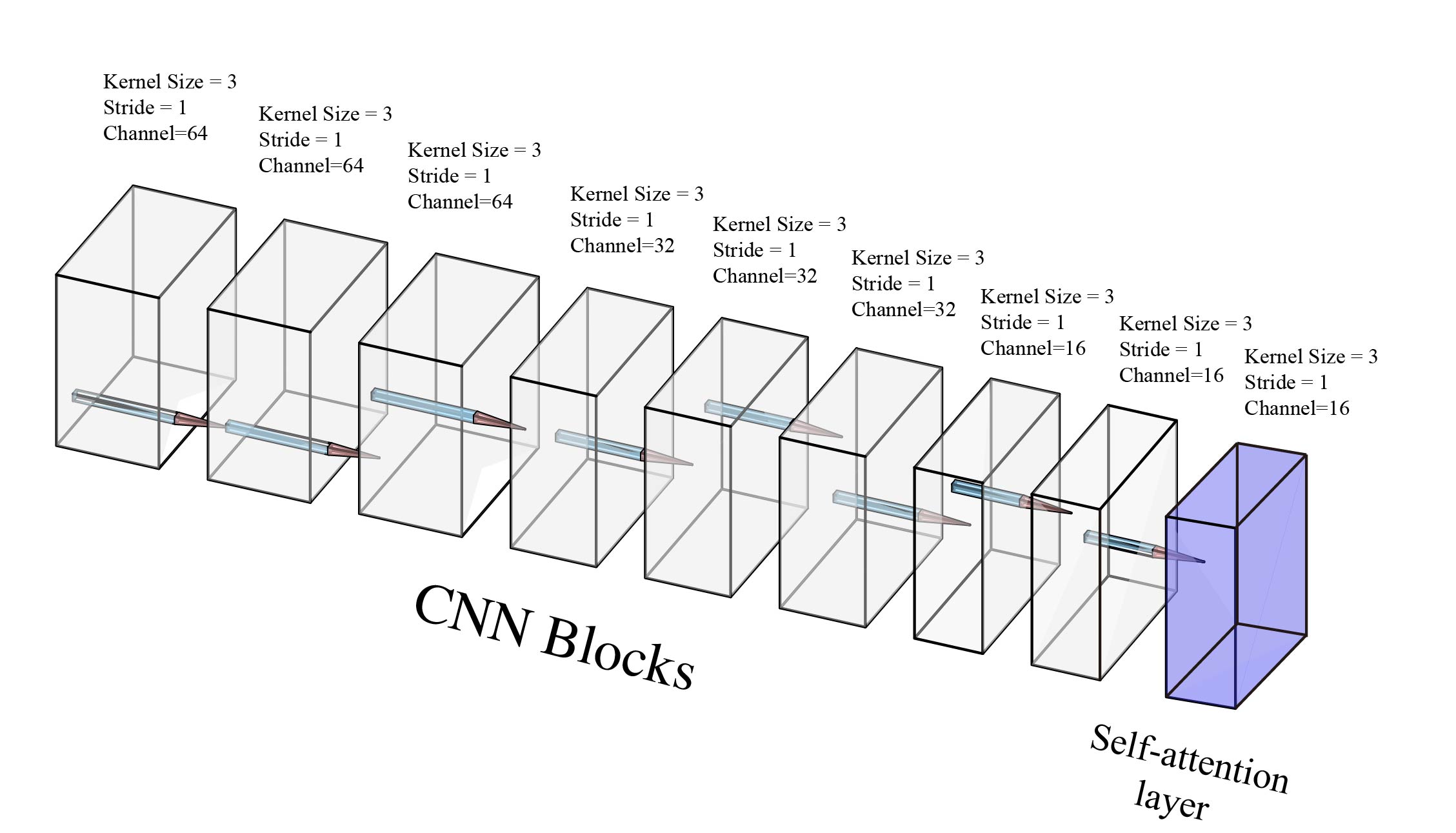}
    \caption{The self-attention based feature extraction block. This block consists of 8 CNN layers and one self-attention layer. The kernel size is set to 3 and the stride is set to 1 for all layers.
    The out channels decrease every 3 layers and the final output channel is 16.}\label{fig:FeatureExtractionBlock}
  \end{center}
\end{figure}

Our feature extraction block consists of 8 convolutional layers and a self-attention layers with 16 output channels,
each of which is followed by a leaky rectified linear unit (Leaky ReLU), as shown in Fig.(\ref{fig:FeatureExtractionBlock}).
In contrast to feature extraction block of previous MVS networks, the self-attention mechanism is introduced
for learning to focus on important information for depth inference.

\begin{figure}[h]
  \begin{center}
    \includegraphics[scale = 0.3]{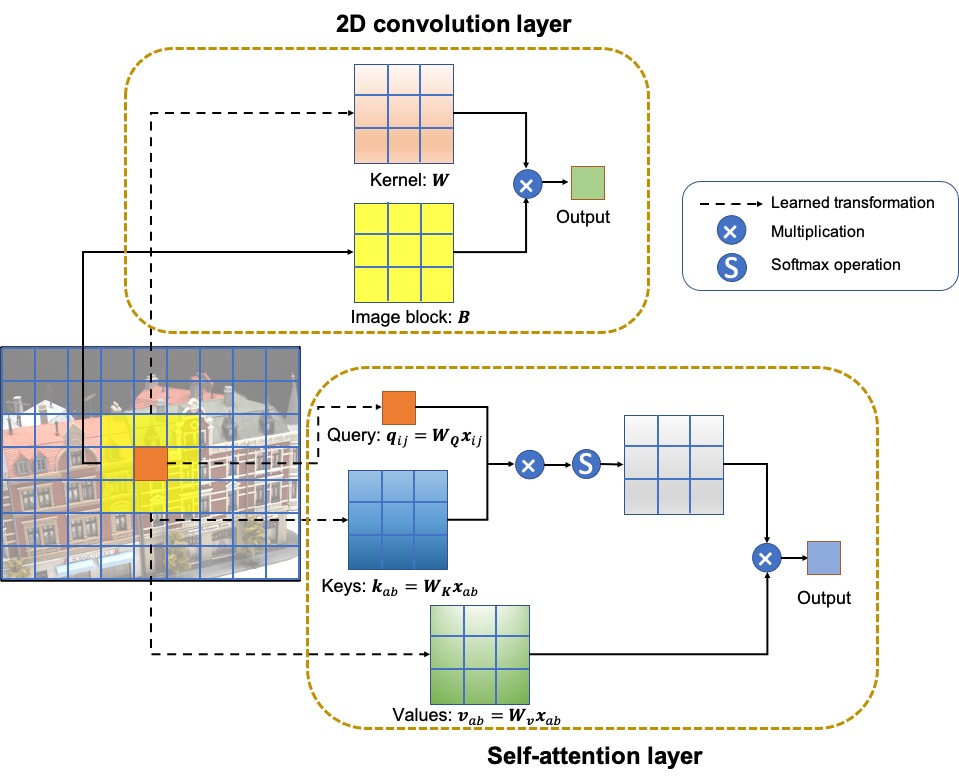}
    \caption{The difference between convolution layer and self-attention layer when the kernel size equals $3$. The convolution 
    layer could be considered as the combination of pixels and learned weights $\mathbf{W}$ while the self-attention layer has three learned weights matrices, namely 
    $\mathbf{W}_Q$, $\mathbf{W}_K$ and $\mathbf{W}_V$. The final output of the self-attention layer is computed through 3 steps. }\label{fig:comparison}
  \end{center}
\end{figure}

Given a learned weight matrix $\mathbf{W} \in \mathbb{R}^{k\times k \times d_{in}\times d_{out}}$
the convolutional output $y_{ij} \in \mathbb{R}^{d_{out}}$ at pixel $(i,j)$ is defined by the summing the linear product of
input image $\mathbf{I}$ and weight matrix:
\begin{equation}
  y_{ij} = \sum_{a,b\in \mathbf{B}}W_{i-a,j-b}\cdot \mathbf{I}(a,b)
  \label{eq:convolution}
\end{equation}
where $k$ denotes kernel size, $d_{in}$ is the input channels quantity while $d_{out}$ is that of output channels,
and $\mathbf{B}$ is the image block for convolution computation with the same size of the kernel.
Compared to traditional attention mechanism \citep{xu2015show,chorowski2015attention,bartunov2018assessing}, the self-attention mechanism is defined as attention applied to a single context instead of across multiple context\citep{ramachandran2019stand}, which
directly models long-distance interactions and leads to state-of-the-art models for various tasks \citep{devlin2018bert, shaw2018self,shazeer2018mesh}.
The self-attention could be formulated as \citep{ramachandran2019stand}
\begin{equation}
  y_{ij} = \sum_{a,b\in \mathbf{B}} \rm{Softmax}_{ab}(\mathbf{q}_{ij}^{\rm{T}}\mathbf{k}_{ab}) \mathbf{v}_{ab}
  \label{eq:selfattentionconvolution}
\end{equation}
where $\mathbf{q}_{ij} = \mathbf{W}_Qx_{ij}$, $\mathbf{k}_{ab} = \mathbf{W}_Kx_{ab}$ and $\mathbf{v}_{ab} = \mathbf{W}_Vx_{ab}$ denote queries, keys and values respectively, and the matrix $\mathbf{W}_{l} (l = Q,K,V)$
consists of the learned parameters. Compared to traditional convolution computation (Eq.(\ref{eq:convolution})), Eq.(\ref{eq:selfattentionconvolution}) could be decomposed into 3 steps:
\begin{itemize}
  \item \textbf{Step 1.} Compute the queries ($\mathbf{q}_{ij}$), keys ($\mathbf{k}_{ab}$) and values ($\mathbf{v}_{ab}$).
  \item \textbf{Step 2.} Measure the similarity of the queries and keys by calculating their inner product $\mathbf{q}_{ij}^{\rm{T}}\mathbf{k}_{ab}$, followed
        by the softmax operation that maps the similarity between $0.0$ and $1.0$.
  \item \textbf{Step 3.} Weight the values with the similarity in \textbf{Step 2} and repeat the all steps for every pixel in $\mathbf{B}$.
\end{itemize}
It is not difficult to figure out that the output $y_{ij}$ is achieved by linear transformations of the pixel
in position $ij$ and the neighborhood pixels, and this operation aggregates spatial information with a convex combination of value vectors with mixing weights parametrized by content interactions.
The difference between the 2D convolution layer and the 2D self-attention layer is
illustrated in Fig.(\ref{fig:comparison}) when $k = 3$.

However, Eq.(\ref{eq:selfattentionconvolution}) does not contain the positional information for queries $\mathbf{q}_{ij}$, which makes it permutation equivariant, limiting expressivity for vision tasks \citep{ramachandran2019stand}. Therefore,
the positional information embedding procedure should be introduced to achieve better results. We introduce the relative position embedding \citep{shaw2018self} rather than the absolute position embedding \citep{vaswani2017attention} methods to bring in
the row and column offset (denoted as $\mathbf{r}_{a-i,b-j}$) into Eq.(\ref{eq:selfattentionconvolution}), and the self-attention computation could be formulated as

\begin{equation}
  y_{ij} = \sum_{a,b\in \mathbf{B}} \rm{Softmax}_{ab}(\mathbf{q}_{ij}^{\rm{T}}\mathbf{k}_{ab}+\mathbf{q}_{ij}^{\rm{T}}\mathbf{r}_{a-i,b-j}) \mathbf{v}_{ab}
  \label{eq:selfattentionconvolutionfinal}
\end{equation}

According to Eq.(\ref{eq:selfattentionconvolutionfinal}), each $y_{ij}$ measures the similarity between the query and an element in $\mathbf{B}$, and enjoys translation equivariance \citep{CordonnierLoukasJaggi2019}.

\subsubsection{Hierarchical feature extraction}
In contrast to previous work such as MVSNet \citep{yao2018mvsnet} that extract features at a fixed scale to generate depth map with a fixed resolution,
our network estimates the depth map by using a coarse-to-fine strategy. Therefore, our feature extraction pipeline consists of two steps.

We denote $l\in \{0,1,\cdots ,L\}$ for image levels from original scales to the coarsest resolution.
The first step is to build a total of $(L+1)$ levels image pyramid $\{\mathbf{I}_i^l\}_{i=1}^N$ for all
the input source images and reference image. The next step is obtaining hierarchical representations at $L$th level
using the feature extraction block presented in Fig.(\ref{fig:FeatureExtractionBlock}). Since the extraction block could be implemented on images captured from different views
and different scales, the learned weights of the block should be shared as shown in Fig.(\ref{fig:NetworkStructure}). We define the extracted feature maps at $l$th level by
$\{\mathbf{f}_{i}^L\}\in \mathbb{R}^{H/2^l\times W/2^l\times Ch}$ for following sections, where $Ch$ denotes the quantity of output channels.

\subsection{Coarse-to-fine depth estimation}
The cost volume is introduced for stereo matching by building the disparity searching space of each pixel with regular
girds \citep{scharstein2002taxonomy,xu2017accurate}. Since the proposed network introduces pyramid structure in image spaces, the Cost Volume Pyramid (CVP) is intuitively
formed. In our proposed network, the CVP is used for depth map inference at coarsest resolution and depth residual estimation at finer scales.

\subsubsection{Depth inference at the coarsest resolution}
We construct the cost volume at coarsest scales as an initial depth map inference. In most learned MVS approaches, the cost volume is generated by transforming all extracted feature maps to the one generated
from the reference image \citep{yao2018mvsnet,yao2019recurrent,gu2020cascade}. We follow these research for cost volume generation at $L$th level while using different feature aggregation method.

Given the depth range $(d_{min},d_{max})$ of the reference image $\mathbf{I}_0^L$, the cost volume is constructed by sampling $M$ fronto-parallel planes uniformly, which could be formulated as
\begin{equation}
  d_m = d_{min} + m (d_{max} - d_{min}) / M
  \label{eq:dcalculation}
\end{equation}
where $m = 0,1,\cdots, M-1$ denotes the hypothesized depth planes. Similar to \cite{yao2019recurrent}, we introduce the differentiable homography matrix $\mathbf{H}^L_i(d)$ for cost volume transformation from the $i$th
source views to reference image at $L$th level, that is,
\begin{equation}
  \mathbf{H}^L_i(d) = \mathbf{K}^L_i \mathbf{R}_i (\mathbf{E} - \frac{\mathbf{t}_0-\mathbf{t}_i}{d}\mathbf{n}_0^T)\mathbf{R}_0^{T}(\mathbf{K}^L_0)^{T}
  \label{eq:homowarping}
\end{equation}
where the upper-case $L$ indicates the level of images and $\mathbf{E}$ denotes the identity matrix.

Eq.(\ref{eq:homowarping}) suggests possible pixel corresponding between feature maps of source views and the reference image.
For $N$ input images, $N$ 4D tensors $\mathbf{f}^L_i$ with the size of $W/2^L \times H/2^L \times M \times Ch$ are generated, which is a memory consuming procedure. Thus, the feature aggregation process
is usually implemented. In contrast to previous learned MVS networks that use the variance based feature aggregation \citep{yao2019recurrent, yang2020cost, gu2020cascade,chen2019point}, we introduce the
average group-wise correlation \citep{Guo2019Group} that build the cost volumes by the similarity measurement for image matching tasks, whose basic idea is splitting
the features into groups and computing correlation maps
group by group. Thus, the first step is to divide the feature channels of feature maps into $G$ groups and
compute the similarity between the $i$th group feature maps between reference image and the $j$th ($j\in\{1,2,\cdots,N-1\}$) wrapped image at hypothesized depth plane $d_m$ as:
\begin{equation}
  \mathbf{S}^{i,L}_{j,d_m} = \frac{1}{Ch/G} \left<\mathbf{f}^{i,L}_{ref}(d_m),\mathbf{f}^{i,L}_{j}(d_m)  \right>
  \label{eq: gwc}
\end{equation}
where $i\in \{0,1,\cdots,G-1\}$, $\left<\cdot,\cdot \right>$ is the inner product, $\mathbf{f}^L_{j}(d_m)$ indicates the interpolated feature map after
wrapping $\mathbf{f}_j^L$ to reference images by using Eq.(\ref{eq:homowarping}), and all operations above
are element-wise. When the feature similarities of all the $G$ groups are computed by Eq.(\ref{eq: gwc}) for $\mathbf{f}^L_{ref}(d_m)$ and $\mathbf{f}^L_{j}(d_m)$,
the original feature maps $\{\mathbf{f}_i^L\}$ could be compressed into a $G$-channel similarity tensor $\mathbf{S}^{L}_{j,d_m}$ of size $G \times H/2^L \times W/2^L$.
After calculating the similarity measurement at all the $M$ hypothesized depth plane, the cost volume
of $j$th source image $\mathbf{C}^L_j = \text{concat}(\mathbf{S}^{L}_{j,0},\mathbf{S}^{L}_{j,1},\cdots, \mathbf{S}^{L}_{j,M-1})$ would be a $G\times H/2^L \times W/2^L \times M$ sized tensor. Thus, the final aggregated cost volume
could be computed as the average similarity of all the views, that is:
\begin{equation}
  \mathbf{C}^L = \frac{1}{N-1}\sum_{j=1}^{N-1}\mathbf{C}^L_j
  \label{eq: cv}
\end{equation}

\begin{figure}[h]
  \begin{center}
    \includegraphics[scale = 0.450]{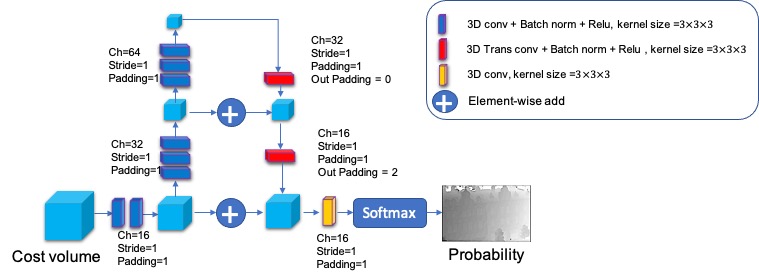}
    \caption{The structure of 3D convolution block. Similar to \cite{yang2020cost}, 
    a multi-scale 3D convolutional network is applied to estimate the probabilities of different depth or residual hypothesis for each pixel.}\label{fig:3DConvBlock}
  \end{center}
\end{figure}

When the cost volume at $L$th level is estimated, the probability volume $\mathbf{P}^L$ could be generated by a 3D convolution block (shown in Fig.(\ref{fig:NetworkStructure}) and Fig.(\ref{fig:3DConvBlock}))
similar to the previous learned MVS networks \citep{gu2020cascade,yang2020cost} and
the depth map of each pixel $\mathbf{p}$ at the coarsest level could be estimated as

\begin{equation}
  \mathbf{D}^L(\mathbf{p}) = \sum_{m=0}^{M-1} d_m \mathbf{P}^L({\mathbf{p}},d_m)
  \label{eq: depth_map}
\end{equation}
where $d_m$ is calculated by using Eq.(\ref{eq:dcalculation}).

\subsubsection{Depth residual estimation at finer scales}
\begin{figure}[h]
  \begin{center}
    \includegraphics[scale = 0.40]{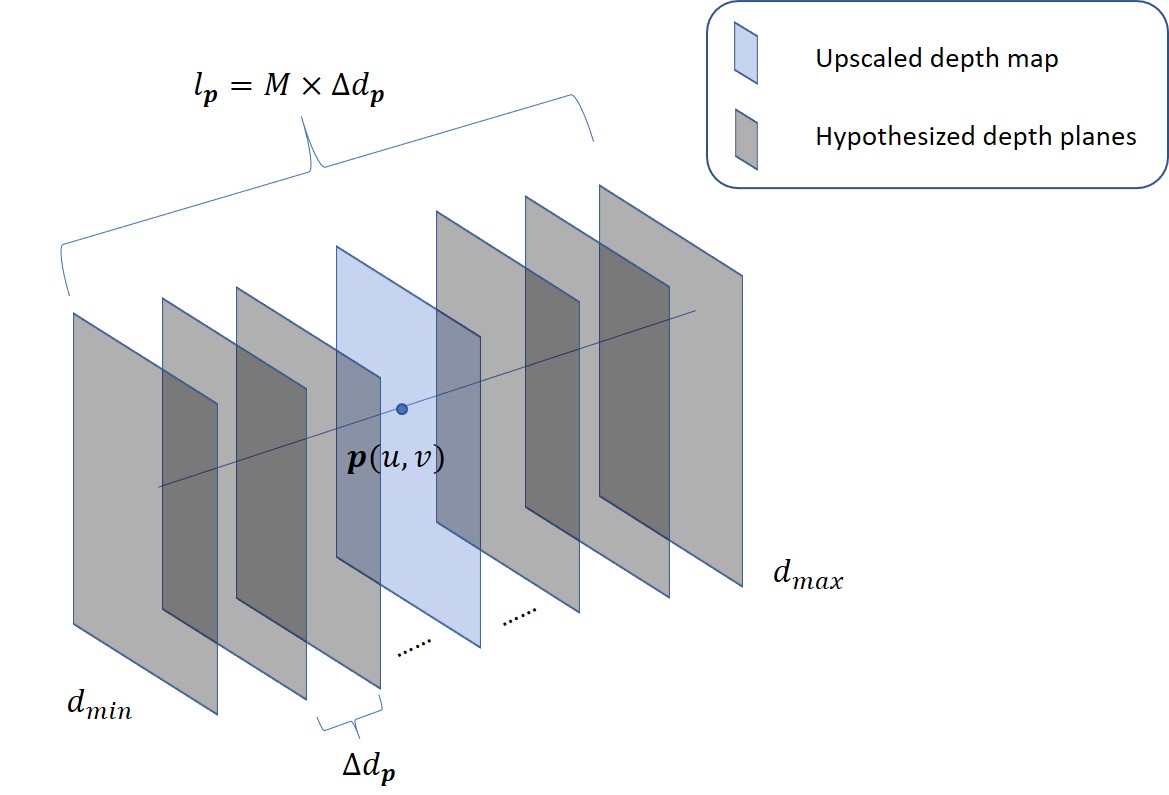}
    \caption{The depth searching range. The upscaled plane is in the middle of searching space, while there are 
    $M/2$ hypothesized planes at each side. $\Delta d_p$ is the depth interval and the total searching distance is $l_{\mathbf{p}}$}\label{fig:HP}
  \end{center}
\end{figure}

Since the depth map $\mathbf{D}^L(\mathbf{p})$ at $L$th level is obtained at the lowest resolution, the quality of 3D reconstruction would be limited. Thus, we try to refine $\mathbf{D}^L(\mathbf{p})$ at
a finer level, and the residual map estimation is intuitively implemented. We take residual map estimation at $(L-1)$th level ($\mathbf{R}^{(L-1)}$) as an example, and the mathematical model could be
summarized as
\begin{equation}
  \left\{
  \begin{array}{lr}
    \begin{split}
      \mathbf{R}^{(L-1)} &= \sum_{m=-M/2}^{M/2}r_{\mathbf{p}}(m)\mathbf{P}_{\mathbf{p}}^{(L-1)}(r_{\mathbf{p}})\\
      \mathbf{D}^{(L-1)}(\mathbf{p}) &=  \mathbf{R}^{(L-1)} + \mathbf{D}_{upscale}^{(L)}(\mathbf{p})
    \end{split}
  \end{array}
  \right.
  \label{eq: residual}
\end{equation}
where $M$ is the number of hypothesized depth planes, $r_{\mathbf{p}} = m \Delta d_{\mathbf{p}}$ represents the depth residual, $\Delta d_{\mathbf{p}} = l_{\mathbf{p}}/M$
is the depth interval, $\mathbf{D}_{upscale}^{(L)}$ is the upscaled depth map from $L$th level and $l_{\mathbf{p}}$ denotes the depth searching range at $\mathbf{p}(u,v)$. As shown in Fig.(\ref{fig:HP}), $\Delta d_{\mathbf{p}}$ and
$l_{\mathbf{p}}$ determine the depth estimation result at each pixel $\mathbf{p}$, and are the key parameters for depth residual estimation.
There exist several methods for depth searching range and depth interval determination, such as iteratively range narrowing \citep{gu2020cascade} and uncertainty based searching boundary \citep{cheng2020deep}.
Here we use the determination method used by \citeauthor{yang2020cost} that projects $\mathbf{p}(u,v)$ in reference image and the corresponding points in source images into
object space and determine depth interval $\Delta d_{\mathbf{p}}$ as the distance of the projection of two neighbor pixels along the epipolar line, because it is pointed by \citeauthor{yang2020cost} that it is not
necessary to sample depth planes densely as the projections of those sampled 3D points in the image are too close to provide extra information for depth inference. Thus, the $l_{\mathbf{p}}$ could be calculated as the
product of $\Delta d_{\mathbf{p}}$ and the given parameter $M$ at level $(L-1)$.

The aggregated cost volume $\mathbf{C}^{(L-1)}$ could be built by using Eq.(\ref{eq: gwc}) and Eq.(\ref{eq: cv}), and depth map  $\mathbf{D}^{(L-1)}$ could be achieved by using Eq.(\ref{eq: residual}) after 3D convolution block
and softmax operation for $\mathbf{P}^{(L-1)}$.

\subsubsection{Iterative depth map estimation}
Recall that our ultimate goal is to estimate $\mathbf{D}^0$ from images at finest level. 
Because our network structure (Fig.(\ref{fig:NetworkStructure})) only uses 2D and 3D convolutional layers
that capture local features and all the weights are shared between different levels, 
we could estimate the depth map iteratively by input $\{\mathbf{I}_i^l\}_{i=0}^{N}(0\leq l < L-1)$ to the
feature extraction block for hierarchical feature maps extraction, 
followed by depth hypothesis estimation with upscaled depth map $\mathbf{D}^{l+1}$ and cost volume generation.
The residual depth $\mathbf{R}^{l}$ could be generated with 3D convolution block and the softmax operation along with the depth map 
$\mathbf{D}^{l}$. Thus, we put $\mathbf{D}^{l}$ as input at $(l-1)$th level and an iterative depth map estimation process is formed. The 
final depth map $\mathbf{D}^0$ will be achieved when we reach the top level in Fig.(\ref{fig:NetworkStructure}).

For network backward propagation, we build the loss function as the sum of $L_1$-norm between ground truth and estimated depth map, that is,

\begin{equation}
  \mathbb{L} = \sum_{l=0}^L \sum_{\mathbf{p}\in \mathbf{\Omega}} \| \mathbf{D}^l(\mathbf{p}) - \mathbf{D}_{GT}^l(\mathbf{p}) \|_1
  \label{eq: loss}
\end{equation}
where $GT$ represents ground truth of depth map and $\mathbf{\Omega}$ is the set of valid pixels. Therefore, the weights could be trained by optimizing Eq.(\ref{eq: loss}) and
the estimated depth map could be achieved by forward propagation using the trained model.
\section{Experiments}
In this section, we present the datasets used in our experiments and training configurations. Afterwards the analysis of the
experiments and ablation studies are presented.

\subsection{Datasets}
We use the DTU dataset, which is now a widely used benchmark for 3D reconstruction, and the recent released BlendedMVS dataset for experiments.
\begin{itemize}
  \item \textbf{DTU dataset}. The DTU dataset consists of 124 different indoor scenes including a variety of objects scanned
        by fixed camera trajectories in 7 different lighting conditions \citep{aanas2016DTU}. The original size of image is $1600\times 1200$ pixels. Following the common configurations, we
        use the same training and evaluation dataset split used by \cite{yao2019recurrent, gu2020cascade, cheng2020deep} and the ground truth
        is provided by \cite{yao2018mvsnet}. This dataset could be used for quantitative analysis.
  \item \textbf{BlendedMVS dataset}. The BlendedMVS dataset is a large-scale MVS dataset for generalized multi-view stereo networks \citep{yao2020blendedmvs}.
        This dataset contains more than 17k MVS training samples covering a variety of 113 scenes, including outdoor buildings, architectures, sculptures and small objects. However, there are no
        official ground truth provided, nor does the evaluation toolbox. Thus, we could only compare results qualitatively.
\end{itemize}

\subsection{Training and evaluation}\label{sec: ExperimentsSettings}

The training of learned MVS methods are usually memory consuming and slow. Since our network is built for depth map estimation iteratively and all the weights are shared between different layers,
we could train our network use the downsampled images and ground truth to boost the training procedure.

The training on DTU dataset was implemented with the image size of $160 \times 128$ pixels along with the corresponding camera parameters including camera intrinsic matrices, rotation matrices and translation vectors provided by \cite{yang2020cost}
while the trained weights were evaluated on full sized images. It is notable that the width and height should be dividable by
$16$ to make the input size suitable for the 3D convolution block (shown in Fig.(\ref{fig:3DConvBlock})). We adopt 3 views for training as is set in \cite{yao2018mvsnet,chen2019point}, and $3$ views are used for
evaluation. More results with different number of views will be shown in Section \ref{sec: nov}.
Both the image pyramid and ground truth pyramid have 2 levels for training and the coarsest resolution of images is $80 \times 64$ pixels. We evaluate our model with a similar size with the training dataset at the coarsest level, so we first crop the images to the resolution
$1600 \times 1184$ pixels and build image pyramid with 5 levels with the resolution $100\times 74$ pixels at the coarsest level.

The training on BlendedMVS dataset was similar to DTU training procedure. Here we use the images with officially provided low resolution ($768\times 576$ pixels) for following experiments.
The images and the ground truth were downsampled to $384 \times 288 $ pixels and the parameters in intrinsic matrices were adjusted correspondingly.
We build image pyramid with 3 levels for training and 4 levels for evaluation, and the coarsest resolution is $96 \times 72$ pixels for each experiment.

In all experiments, we set the number of hypothesized depth plane $M=48$ at coarsest level and $M=8$ at other levels for both training and evaluation. The networks are trained on 4 Nvidia GeForce RTX 2080Ti graphics cards for 40 epochs with batch size set to 36 (mini-batch size of 9 per GPU).
We used Adam \citep{2014Adam} to optimize the proposed network and the initial learning rate is set to $1 \times 10^{-3}$ which is multiplied by $0.5$ at 10th, 25th, 32nd
epoch.

\subsection{Post processing and accuracy evaluation}
We fuse all the depth maps into a complete one and generate the dense point cloud by the Fusibile toolbox \citep{2015Massively} as used in \cite{gu2020cascade, yang2020cost, cheng2020deep},  which consists of
three steps for point cloud generation: photometric filtering, geometric consistency filtering, and depth fusion.

To quantitatively evaluate the 3D-reconstruction performance on DTU dataset,
we calculate both the mean accuracy (referred as Acc.), the mean completeness (referred as Comp.) and the overall accuracy (referred as OA) which is defined as:
\begin{equation}
  OA = \frac{Acc. + Comp.}{2}
  \label{eq: oa}
\end{equation}
The accuracy and completeness could be calculated using the official MATLAB script provided by the DTU dataset \citep{aanas2016DTU}.
\subsection{Results on DTU dataset}
We first compare our results to those reported by traditional geometric-based methods and other learning-based baseline methods. As summarized in Tab.(\ref{tab:OA}),
our method outperforms all methods in terms of completeness and overall accuracy with the number of group is set to $G = 4$, while Gipuma \citep{2015Massively} performs the best in terms of accuracy.
Here we found an interesting phenomenon that the number of group is a quarter of original channel quantity which is the same as channel deduction implemented in
the MVSNet that reduces the 32-channel cost volume to an 8-channel one, before taken into 3D regularization block. Meanwhile, it is also demonstrated by \cite{tulyakov2018practical} that
a compressed 8-channel cost volume could achieve similar accuracy compared with original 32-channel cost volume in image matching task. This makes us believe that the raw cost volume representation may be redundant, and the cost volume could be compressed to the one
with fewer channel quantity. Whether a quarter of original channel quantity is the best choice should be proved theoretically or validated with more experiments.

\begin{figure}[h]
  \begin{center}
    \includegraphics[scale = 0.23]{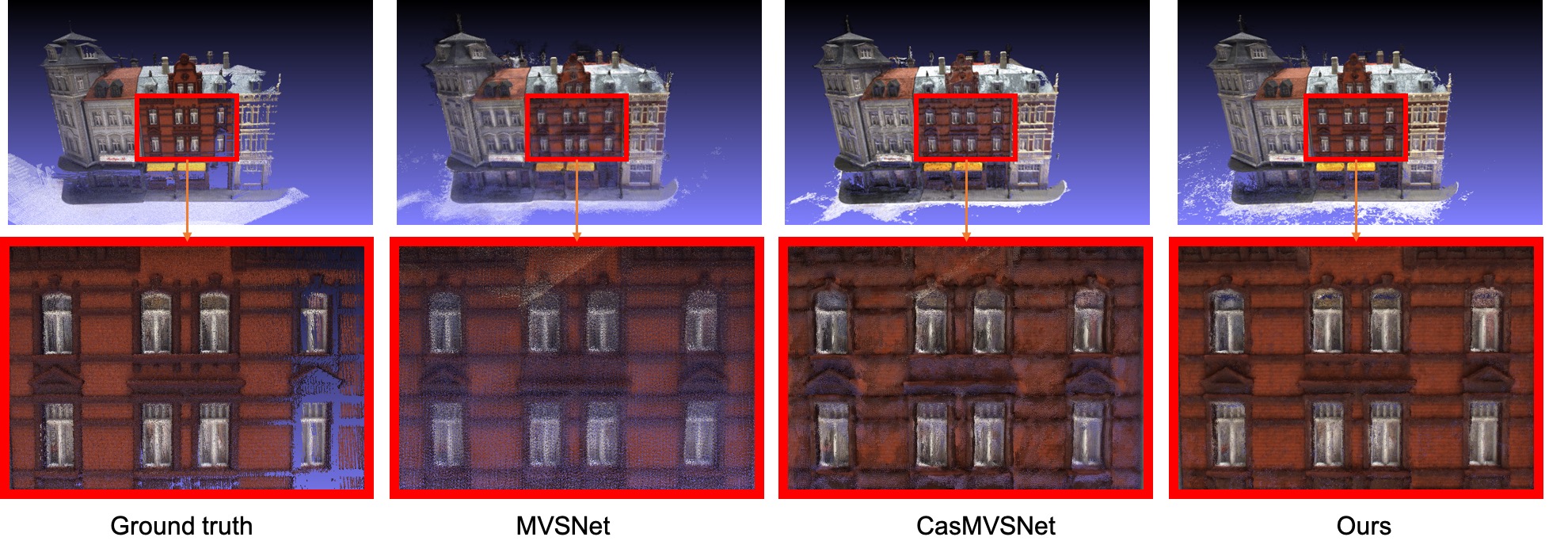}
    \caption{3D reconstruction result of 9th scene in DTU dataset.}\label{fig:ResultDTU9}
  \end{center}
\end{figure}

\begin{figure}[h]
  \begin{center}
    \includegraphics[scale = 0.20]{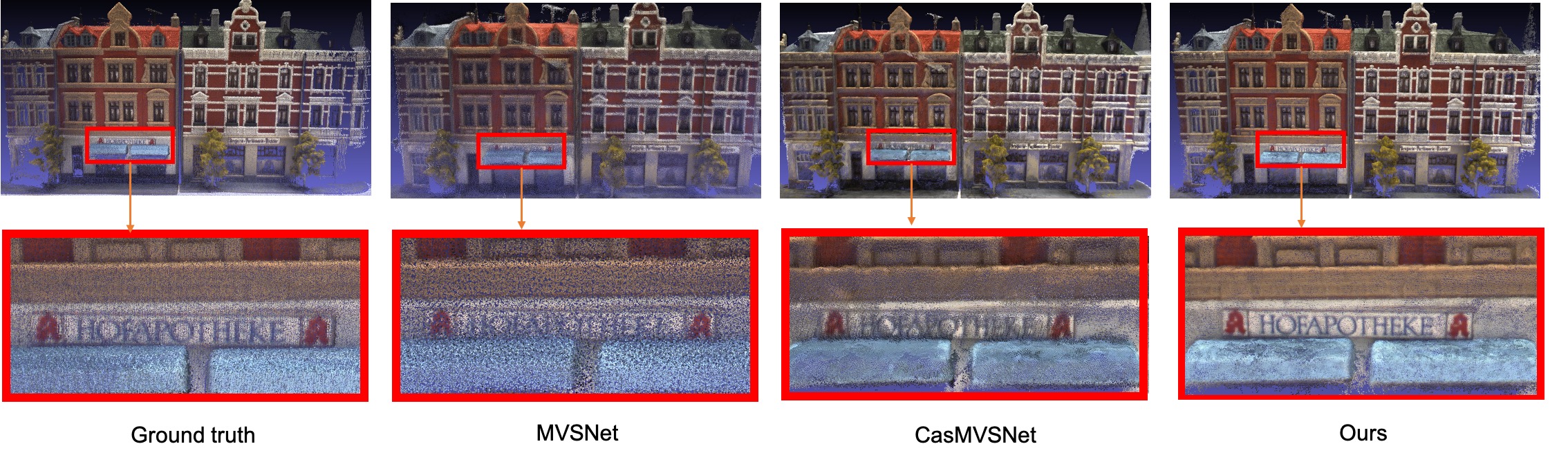}
    \caption{3D reconstruction result of 15th scene in DTU dataset.}\label{fig:ResultDTU15}
  \end{center}
\end{figure}

\begin{figure}[h]
  \begin{center}
    \includegraphics[scale = 0.25]{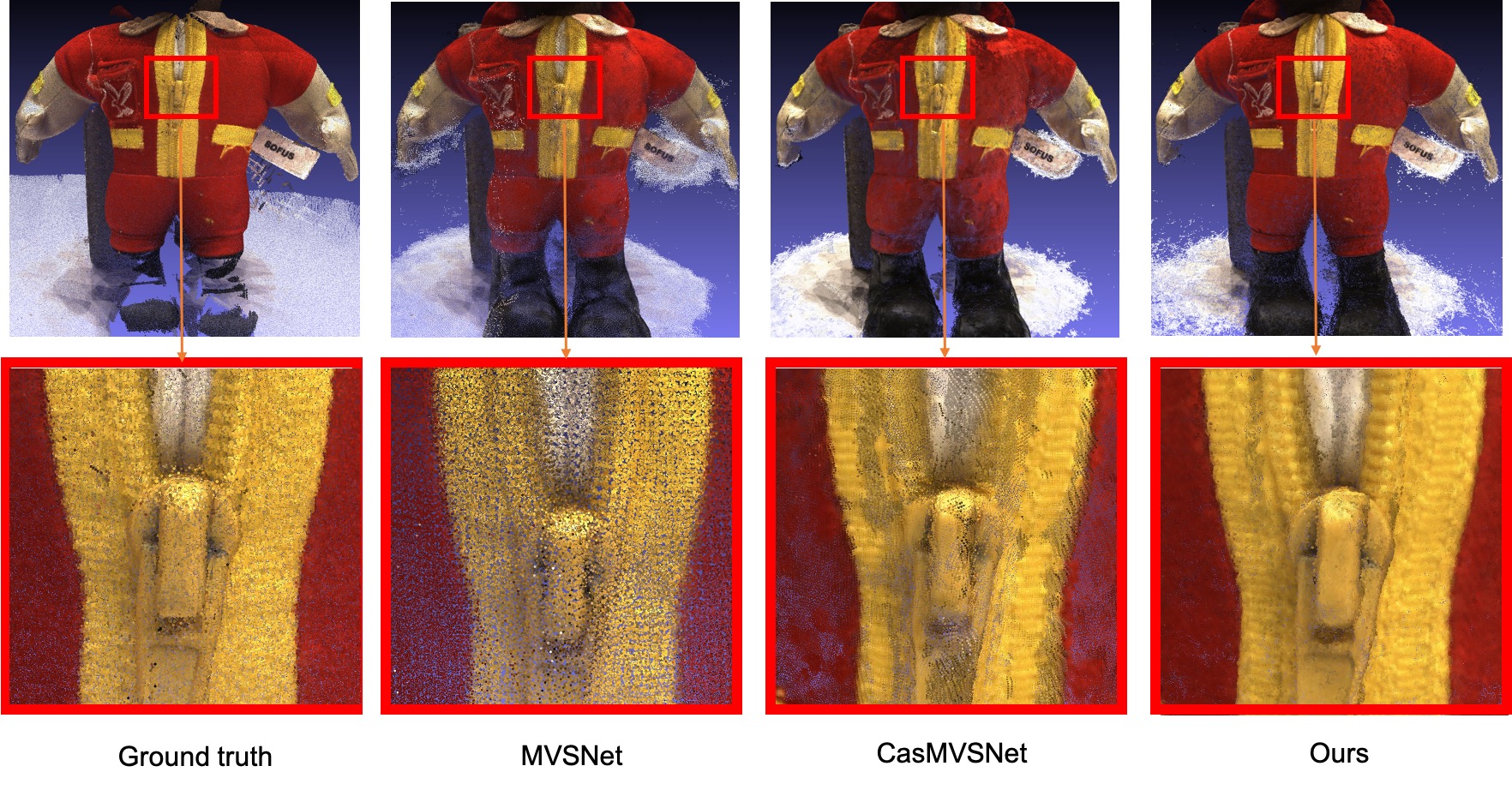}
    \caption{3D reconstruction result of 49th scene in DTU dataset.}\label{fig:ResultDTU49}
  \end{center}
\end{figure}

We compare
our 3D reconstruction results with MVSNet and CasMVSNet in Fig.(\ref{fig:ResultDTU9}), Fig.(\ref{fig:ResultDTU15}) and Fig.(\ref{fig:ResultDTU49}). Both the AACVP-MVSNet and
CasMVSNet achieve comparable completeness in these examples while the point cloud generated by MVSNet is sparser as they were extracted with low resolution depth maps. 
As we can see in Fig.(\ref{fig:ResultDTU9}) and Fig.(\ref{fig:ResultDTU49}), our results are smoother on the surfaces. Moreover, the letters in Fig.(\ref{fig:ResultDTU15}) is more complete and
could be more easily recognized from 3D reconstruction results.

We also show the memory usage by AACVP-MVSNet in Tab.(\ref{tab:MemoryUsage}), and the baseline is the network whose cost volumes were generated by variance based method used in \cite{yao2018mvsnet, chen2019point, xiang2020pruning, gu2020cascade} instead of the similarity measurement based method
used in our network while the other
parts stay the same with Fig.(\ref{fig:NetworkStructure}). As indicated in Tab.(\ref{tab:MemoryUsage}), the memory usage is around $10\%$ less than the baseline, and the memory usage decreases slightly with
as the parameter $G$ decreases, which could be easily illustrated by Eq.(\ref{eq: gwc}) and Eq.(\ref{eq: cv}).

\begin{table}[htbp]
  \caption{\label{tab:OA} Quantitative results of reconstruction quality on the
    DTU evaluation dataset (lower is better).}
  \centering
  \begin{tabular}{lccc}
    \toprule
    Methods                                      & Acc.(mm)       & Comp. (mm)     & OA (mm)        \\
    \midrule
    Camp \citep{campbell2008using}               & 0.835          & 0.554          & 0.695          \\
    Gipuma  \citep{2015Massively}                & \textbf{0.283} & 0.873          & 0.578          \\
    Furu  \citep{furukawa2009accurate}           & 0.613          & 0.941          & 0.777          \\
    \midrule
    SurfaceNet \citep{ji2017surfacenet}          & 0.450          & 1.040          & 0.745          \\
    $\text{MVSNet}^*$ \citep{yao2018mvsnet}      & 0.396          & 0.527          & 0.462          \\
    $\text{R-MVSNet}^*$ \citep{yao2019recurrent} & 0.383          & 0.452          & 0.418          \\
    PruMVSNet \citep{xiang2020pruning}           & 0.495          & 0.433          & 0.464          \\
    PointMVSNet \citep{chen2019point}            & 0.361          & 0.421          & 0.391          \\
    CasMVSNet \citep{gu2020cascade}              & 0.325          & 0.385          & 0.355          \\
    CVP-MVSNet \citep{yang2020cost}              & 0.296          & 0.406          & 0.351          \\
    Ours($G = 8$)                                & 0.363          & 0.332          & 0.347          \\
    Ours($G = 2$)                                & 0.360          & 0.341          & 0.351          \\
    \midrule
    Ours($G = 4$)                                & 0.357          & \textbf{0.326} & \textbf{0.341} \\
    \bottomrule
  \end{tabular}

  *Official MVSNet and R-MVSNet implementation used the Altizure internal library for post-processing and could achieve
  higher accuracy compared to the Fusibile toolbox.
\end{table}

\begin{table}[htbp]
  \caption{\label{tab:MemoryUsage} Memory used per batch by AACVP-MVSNet with different parameter $G$. The
    training images were resized to $160 \times 128$ pixels.}
  \centering
  \begin{tabular}{lccccc}
    \toprule
    \quad                & Baseline            & $G = 4$                      & $G = 8$             & $G=16$              \\
    \midrule
    Memory usage (Mb)         & $1171.44$ & $\mathbf{1048.44}$ & $1065.89$ & $1087.93 $ \\
    Compared to baseline (\%) & $-0$              & $-\mathbf{10.50}$          & $-9.01 $          & $-7.13$           \\
    \bottomrule
  \end{tabular}
\end{table}

\subsection{Results on BlendedMVS dataset}
Since the BlendedMVS dataset does not contain officially provided 3D reconstruction results, nor does any SOTA algorithms provide officially BlendedMVS dataset training implementation
except MVSNet and R-MVSNet that only provide pre-trained weights without fusion results,
we only show some qualitative results. We
first pick up all the large scale outdoor scenes from all samples, and split the picked scenes with the ratio $4:1$ for training and evaluation.

\begin{figure}[htbp]
  \centering

  \subfigure[Outdoor Scene1.]{
    \begin{minipage}[t]{0.5\linewidth}
      \centering
      \includegraphics[scale=0.2]{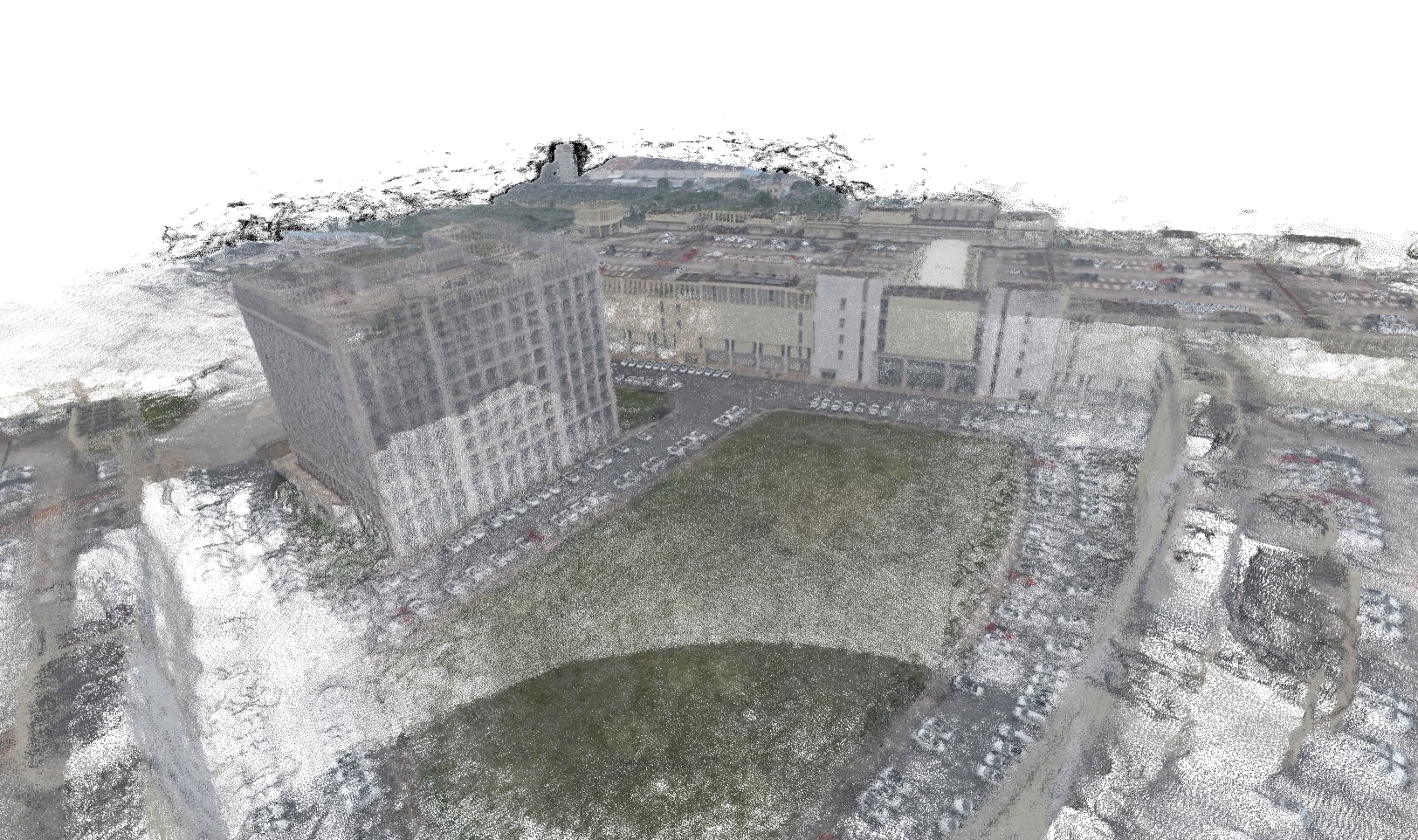}
    \end{minipage}%
  }%
  \subfigure[Outdoor Scene2.]{
    \begin{minipage}[t]{0.5\linewidth}
      \centering
      \includegraphics[scale=0.2]{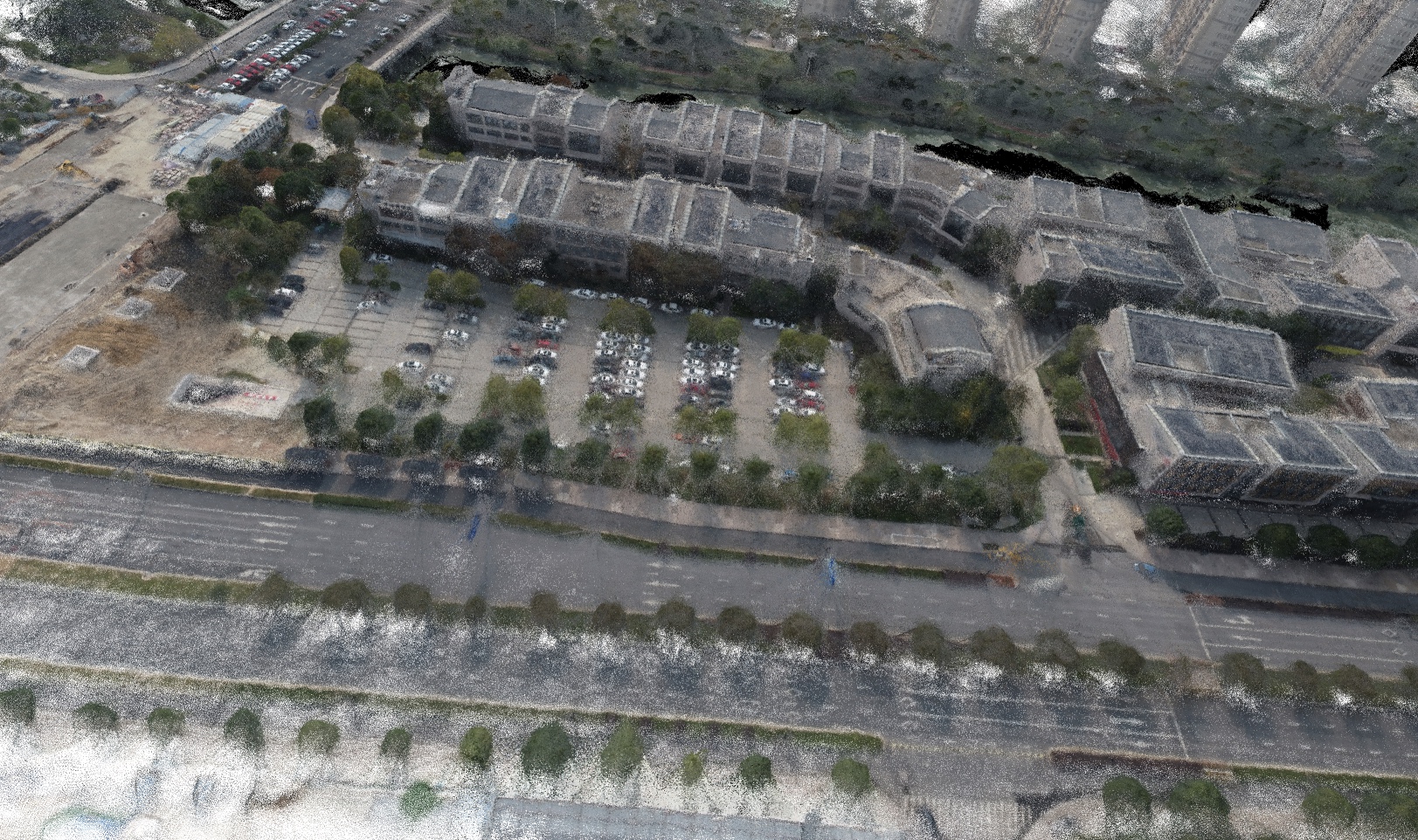}
    \end{minipage}%
  }%
  \quad
  \subfigure[Outdoor Scene3.]{
    \begin{minipage}[t]{0.5\linewidth}
      \centering
      \includegraphics[scale=0.2]{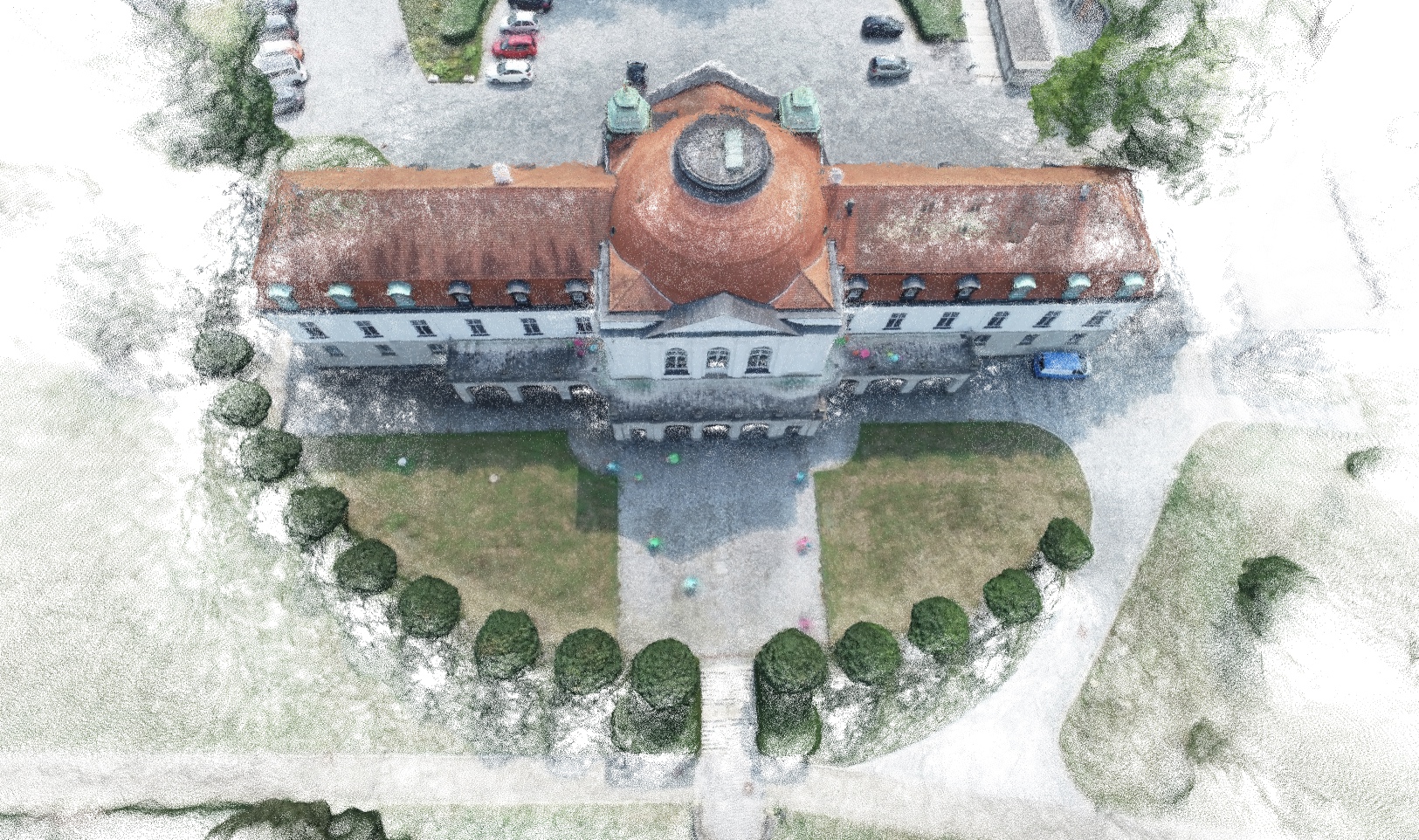}
    \end{minipage}
  }%
  \subfigure[Outdoor Scene4.]{
    \begin{minipage}[t]{0.5\linewidth}
      \centering
      \includegraphics[scale=0.2]{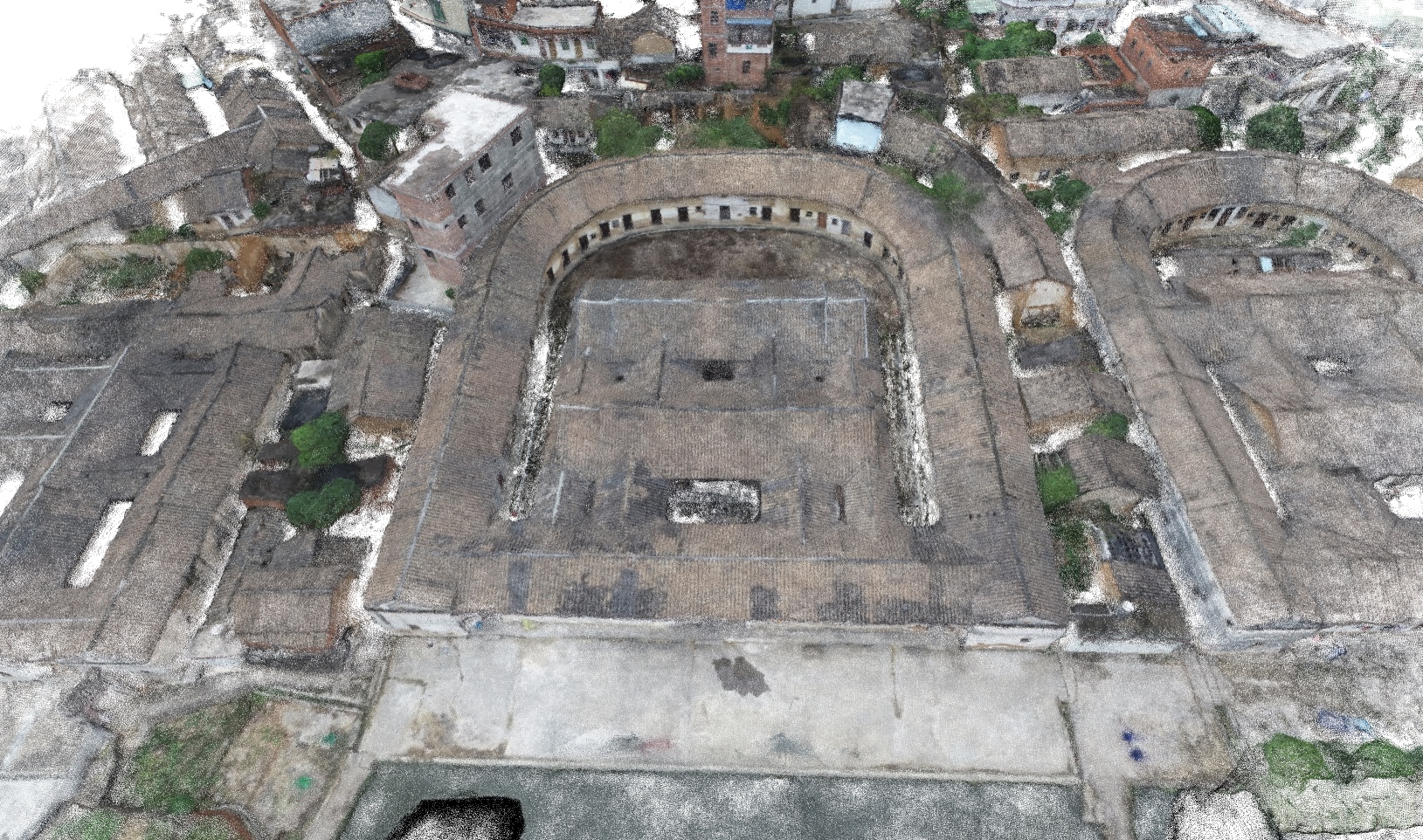}
    \end{minipage}
  }%

  \centering
  \caption{ Results on the BlendedMVS dataset.} \label{fig: ResultBMVS}
\end{figure}

\begin{figure}[h]
  \begin{center}
    \includegraphics[scale = 0.15]{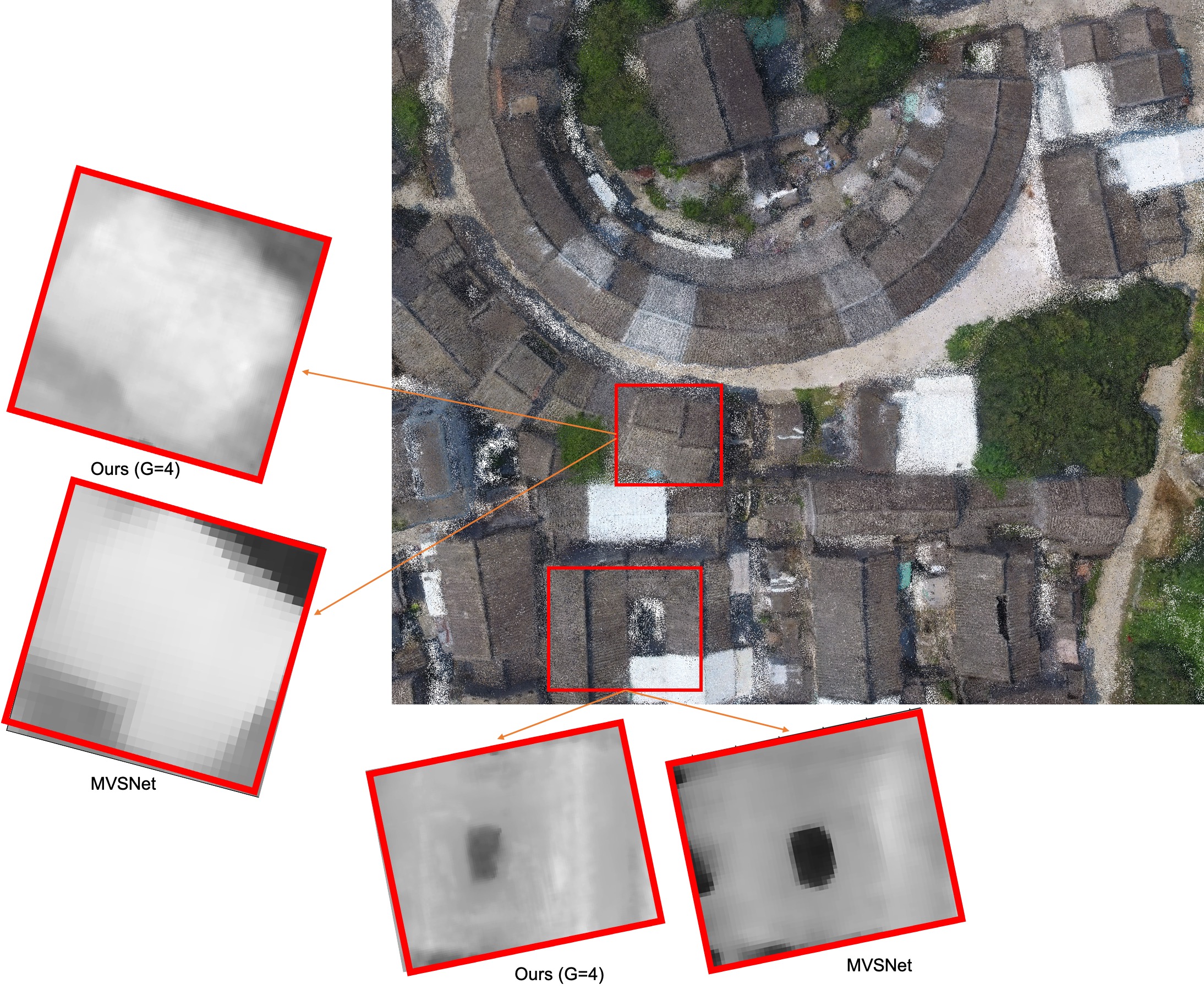}
    \caption{Comparison of depth inference results between MVSNet and AACVP-MVSNet. }\label{fig:ResultBMVS1}
  \end{center}
\end{figure}

We show some 3D reconstruction results in Fig.(\ref{fig: ResultBMVS}), and it is easily known that our generated point clouds are smooth and complete.
The comparison of depth map generation results between AACVP-MVSNet and MVSNet is shown in Fig.(\ref{fig:ResultBMVS1}), and it's obvious that our depth map has a higher resolution and
while captures more high-frequency details in edgy areas.

\subsection{Ablation Studies}
In this section, we provide ablation experiments and quantitative analysis to evaluate the strengths and limitations of the key
components in our framework.
For all the following studies, experiments are performed and evaluated on the DTU dataset,
and both accuracy and completeness are used to measure the reconstruction quality. We set the number of groups $G = 4$, and all
settings are the same as used in Section \ref{sec: ExperimentsSettings}.
\subsubsection{Multi-head self-attention layers}
In practice, multiple attention heads are used to learn multiple distinct representations of the input in many applications. It could be
implemented by partitioning the input $x_{ij}$ in Eq.(\ref{eq:selfattentionconvolution}) and Eq.(\ref{eq:selfattentionconvolutionfinal}) into
depth-wise into $H$ groups $x_{ij}^h \in \mathbb{R}^{d_{in}/H} (h = 0,1,\cdots,H-1)$ (the head), and compute the learned parameters $\mathbf{W}^h_{l} (l = Q,K,V)$
for all the heads, followed by concatenating the output representations into $y_{ij}$ as output \citep{ramachandran2019stand}.

We set $H = 2,4,8$ respectively, and the reconstruction quality stays almost the same when $H=2,4$, as shown in Tab.(\ref{tab:ASHeads}) where the single-headed denotes the
self-attention layer used in Fig.(\ref{fig:FeatureExtractionBlock}).
When $H$ is set to $8$, the overall accuracy of reconstruction decreases slightly, which illustrates that the number of channels may not be
too few in each group ($2$ channels in a group when $H = 8$).

\begin{table}[htbp]
  \caption{\label{tab:ASHeads} Reconstruction quality on DTU dataset by AACVP-MVSNet with different parameter $H$.}
  \centering
  \begin{tabular}{lccccc}
    \toprule
    \quad      & Singe-headed     & $H = 2$          & $H = 4$ & $H=8$   \\
    \midrule
    Acc. (mm)  & $\textbf{0.357}$ & $0.362$          & $0.359$ & $0.375$ \\
    Comp. (mm) & $0.326$          & $\textbf{0.325}$ & $0.332$ & $0.339$ \\
    OA (mm)    & $\textbf{0.341}$ & $0.343$          & $0.345$ & $0.357$ \\
    \bottomrule
  \end{tabular}
\end{table}

\subsubsection{Number of views in training and evaluation} \label{sec: nov}
Since multi-view images would provide more information for depth inference task, we choose the number of views for training $nViews_{T} = 3,5,7$ and
the number of views for evaluation $nViews_{E} = 3,4,5$ for experiments. Tab.(\ref{tab:nviews}) shows the comparison results, which illustrates that the result of 
reconstruction would improve when the number of views for evaluation increases while the accuracy stays almost the same with variant of views quantity
for training. The best overall accuracy in our experiments is $0.326 \text{mm}$ when $nViews_{T} = 7$ and $nViews_{E} = 5$, which is the best result on 
DTU dataset so far as we know.
\begin{table}[htbp]
  \caption{\label{tab:nviews} Reconstruction quality on DTU dataset with different number of views.}
  \centering
  \begin{tabular}{lccc}
    \toprule
    \quad                           & Acc. (mm)        & Comp. (mm)       & OA (mm)          \\
    \midrule
    $nViews_{T} = 3, nViews_{E} =3$ & ${0.357}$        & $0.326$          & $0.341$          \\
    $nViews_{T} = 3, nViews_{E} =4$ & ${0.355}$        & $0.320$          & $0.338$          \\
    $nViews_{T} = 3, nViews_{E} =5$ & ${0.359}$        & $0.319$          & $0.339$          \\
    $nViews_{T} = 5, nViews_{E} =3$ & $0.361$          & $0.326$          & $0.343$          \\
    $nViews_{T} = 5, nViews_{E} =4$ & $0.363$          & ${0.309}$        & ${0.336}$        \\
    $nViews_{T} = 5, nViews_{E} =5$ & $0.365$          & ${0.305}$        & ${0.335}$        \\
    $nViews_{T} = 7, nViews_{E} =3$ & $0.361$          & ${0.326}$        & ${0.343}$        \\
    $nViews_{T} = 7, nViews_{E} =4$ & $\textbf{0.349}$ & ${0.306}$ & ${0.328}$ \\
    $nViews_{T} = 7, nViews_{E} =5$ & ${0.353}$ & $\textbf{0.299}$ & $\textbf{0.326}$ \\
    \bottomrule
  \end{tabular}
\end{table}
\subsubsection{Convergence}
\begin{figure}[h]
  \begin{center}
    \includegraphics[scale = 0.25]{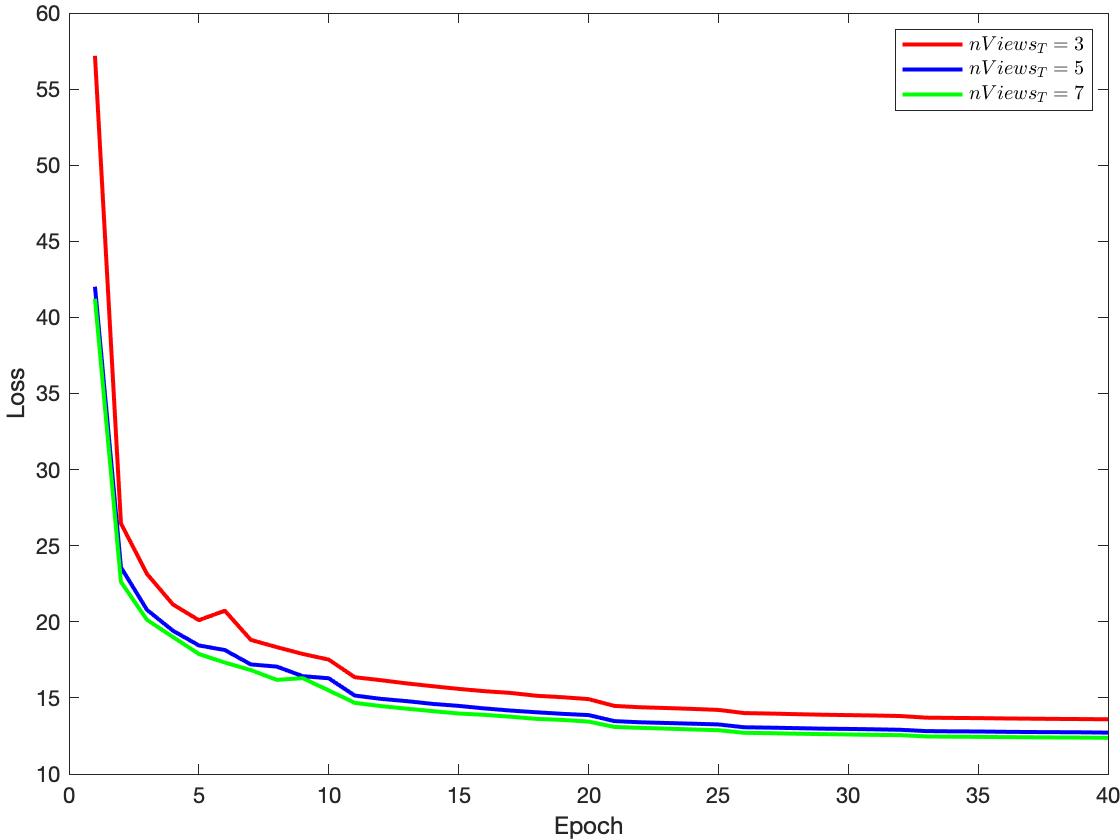}
    \caption{Training loss with $nViews_{T} = 3,5,7$. }\label{fig:T-L}
  \end{center}
\end{figure}

The train-loss curves is plotted in Fig.(\ref{fig:T-L}) with $nViews_{T} = 3,5,7$, which illustrates that our model could converge and 
achieve similar loss at $40\text{th}$ epoch.

\section{Discussion}
Based on the results shown above, we could confirm that the pro­posed architecture, 
which take advantages of self-attention and similarity measurement based cost volume generation in 3D reconstruction,
could be trained iteratively with a coarse-to-fine strategy and achieves better performance than some state-of-the-art methods by extensive evaluation on benchmark datasets.
In this section, we discuss the limitations of our work and some possible future works.

\subsection{Depth searching range at finer levels}
The depth searching range in AACVP-MVSNet is determined by the distance of the projection of two neighbor pixels along
the epipolar lines in images at finer levels in our architecture. Though it is tested in \cite{yang2020cost} that evaluated the 3D reconstruction accuracy
as a function of the interval setting, this method is still an empirical way for depth search space determination. In the future, we will investigate whether the depth searching
space could be learned in the network.

\subsection{Number of hypothesized depth planes}
Both in our work and some previous learned MVS methods such as \citealp{chen2019point, yang2020cost}, the number
of hypothesized depth planes $M$ is determined empirically. Thus, the interval of hypothesized planes might not be
the most reasonable ones for depth map estimation at each level, which may limits the usage of the proposed architecture
in different scenes. We would like to research and propose a self-adaptive method for the determination of the parameter $M$ in the future.

\subsection{Learned MVS methods without ground truth}
As to our knowledge, most MVS networks are based on supervised learning that requires ground truth which may result in lack of accuracy when the scenes are not
similar to the dataset that used for achieving the pre-trained weights. Though some unsupervised MVS methods are proposed recently \citep{dai2019mvs2,Huang2020M3VSNet}, the accuracy and completeness are still
far lower than those of the supervised methods. In the future, we will research on this topic and extend the application of MVS networks to more situations.

\section{Conclusion}
In this paper, we proposed AACVP-MVSNet for MVS problems, which is equipped with the self-attention based
feature extraction and similarity measurement based cost volume
generation method. The AACVP-MVSNet can estimate the depth map by using a coarse-to-fine strategy. The experimental results show that AACVP-MVSNet outperforms some state-of-the-arts MVS networks after an extensive evaluation 
on two challenging benchmark datasets. 
In the future, we want to further improve our network, especially replacing those empirical parameters with the self-adaptive ones.

\section{Acknowledgment}
This work is supported by National Natural Science Foundation of China (No.41801388 and No.41801319).

\bibliography{mybibfile}

\end{document}